\pdfoutput=1
\documentclass{article}

\usepackage{ijcai25}

\usepackage{times}
\usepackage{soul}
\usepackage{url}
\usepackage[hidelinks]{hyperref}
\usepackage[utf8]{inputenc}
\usepackage[small]{caption}
\usepackage{graphicx}
\usepackage{amsmath}
\usepackage{amsthm}
\usepackage{booktabs}
\usepackage{algorithm}
\usepackage{algorithmic}
\usepackage[switch]{lineno}
\usepackage{xspace}
\usepackage{multirow}
\usepackage{makecell}
\usepackage{xspace}
\usepackage{amsmath,mathtools}
\usepackage{color}
\usepackage{xcolor}
\usepackage{bbding}
\usepackage{pifont}
\usepackage{wasysym}
\usepackage{amssymb}

\definecolor{todocolor}{rgb}{0.9,0.1,0.1}
\definecolor{hycolor}{rgb}{0.7,0.7,0.3}

\newcommand{\hwrevised}[1]{\textcolor{black}{#1}}

\newcommand{\modelname}{\textsc{MSViT}\xspace}

\title{\modelname: Improving Spiking Vision Transformer Using Multi-scale Attention Fusion }

\author{
Wei Hua$^1$
\and
Chenlin Zhou$^2$
\and
Jibin Wu$^{3}$
\and
Yansong Chua$^{1*}$\And
Yangyang Shu$^4 $\thanks{Corresponding authors.}\\
\affiliations
$^1$China Nanhu Academy of Electronics and Information Technology, China\\
$^2$University of Chinese Academy of Sciences, China\\
$^3$The Hong Kong Polytechnic University, Hong Kong SAR, China\\
$^4$School of Systems and Computing, The University of New South Wales, Australia\\
\emails
\{huawei, caiyansong\}@cnaeit.com,
zhouchenlin19@mails.ucas.ac.cn,
jibin.wu@polyu.edu.hk,
yangyang.shu@unsw.edu.au
}

\begin{document}

\maketitle

\begin{abstract}
The combination of Spiking Neural Networks (SNNs) with Vision Transformer architectures has garnered significant attention due to their potential for energy-efficient and high-performance computing paradigms.
However, a substantial performance gap still exists between SNN-based and ANN-based transformer architectures.  While existing methods propose spiking self-attention mechanisms that are successfully combined with SNNs, the overall architectures proposed by these methods suffer from a bottleneck in effectively extracting features from different image scales. In this paper, we address this issue and propose \modelname. This novel spike-driven Transformer architecture firstly uses multi-scale spiking attention (MSSA) to enhance the capabilities of spiking attention blocks.
We validate our approach across various main data sets. The experimental results show that \modelname outperforms existing  SNN-based models, positioning itself as a state-of-the-art solution among SNN-transformer architectures. The codes are available at
{\url {https://github.com/Nanhu-AI-Lab/MSViT}}.

\end{abstract}

\section{Introduction}



 Spiking Neural Networks (SNNs), referred to as third-generation neural networks \cite{maass1997networks}, have garnered significant attention \hwrevised{attributed to} their biological plausibility, event-driven processing, and potential for high energy efficiency \cite{roy2019towards,pei2019towards}. Despite these advantages, SNNs have yet to achieve performance levels comparable to traditional Artificial Neural Networks (ANNs), particularly in complex vision tasks. This performance gap presents a major obstacle to the widespread adoption of SNNs in practical applications.


Initially developed for natural language processing tasks, Transformers \cite{vaswani2017attention} have been extensively explored and \hwrevised{extended} to various computer vision applications, including image classification \cite{dosovitskiy2020image}, object detection \cite{liu2021swin}, and semantic segmentation \cite{wang2021pyramid}.  Notably, the self-attention mechanism—the core component of Transformers enables the models to focus on salient input information selectively, bears a striking resemblance to the selective attention processes discovered in the human biological system \cite{whittington2018generalisation}.

Given the biological plausibility of self-attention mechanisms and their alignment with human cognitive processes, it is intuitive to investigate their integration within Spiking Neural Networks (SNNs) to enhance deep learning models. The combination of the powerful representational capabilities of Transformers with the energy-efficient nature of SNNs presents an exciting research direction. In recent years, several studies have explored incorporating spiking neurons into Transformer-based architectures. For instance, Zhou et al. \cite{zhou2022spikformer} introduced "Spikformer," a spike-driven self-attention mechanism, marking the first attempt to integrate spike-driven neurons into the Transformer framework. Shi et al. \cite{shi2024spikingresformer} proposed SpikingResformer, which combines ResNet-inspired architecture with self-attention computation to achieve competitive performance in image classification tasks while maintaining low energy consumption. Yao et al. \cite{yao2023spike,yao2024spike-v2} developed two versions of Spike-driven Transformer trained using sparse AND-ACcumulate (AC) operations, achieving state-of-the-art (SOTA) results on the ImageNet-1K dataset among SNN-based architectures in the same terms.

Despite these advancements, there remains a performance gap between SNN-based models and traditional Artificial Neural Network (ANN) counterparts. While spiking neurons offer energy-efficient processing, their binary nature (using only 0 and 1 spikes) poses significant challenges in training larger and deeper networks. Unlike ANNs, which rely on floating-point matrix multiplication and softmax operations, the binary representation in SNNs struggles to capture the intrinsic and diverse information present in input data, often resulting in suboptimal accuracy for downstream tasks. Addressing these limitations is critical to exploring the full potential of SNN-based Transformer architectures.

Multi-scale structures are extensively utilized across computer vision (CV) \cite{fan2021multiscale}, natural language processing (NLP) \cite{guo2020multi}, and signal processing domains \cite{park2019specaugment} due to their effectiveness in capturing patterns at varying scales. Studies on the visual cortex of cats and monkeys suggest that increasing the number of distinct channels, with each channel corresponding to progressively specialized features, enhances the model’s ability to extract structured information from input images \cite{fan2021multiscale,koenderink1984structure}. These findings demonstrate that multi-scale feature extraction improves the robustness and generalizability of learned representations in CV tasks.


Motivated by these insights, we propose a novel spiking attention mechanism, MSSA, which incorporates \underline{M}ulti-\underline{S}cale \underline{S}pike \underline{A}ttention (MSSA) blocks into transformer architecture. Each attention head within the MSSA block operates at varying scales via multiple inputs, thereby enriching the perceptual field of the spiking self-attention mechanism. Larger scales capture more global and smoother features, while smaller scales focus on local details, enhancing the sharpness and distinctiveness of feature representations. The balance between global and local information of inputs is crucial for spiking neural networks (SNNs), where it is essential to develop innovative methods that expand the representational capacity of spike-based models.

Building on MSSA, we further introduce the Multi-Scale Spike-driven Transformer, comprising hierarchical layers with MSSAs. Each layer is designed with attention heads that process input images at different scales, enabling the network to capture multi-scale features effectively. This architecture bridges the seminal concept of multi-scale feature hierarchies with spike-driven Transformer models, leveraging principles of resolution and channel scaling to improve performance in various visual recognition tasks. We hypothesize that integrating multi-scale attention mechanisms into spike-driven Transformers will significantly enhance their capability to extract diverse and robust features, thereby advancing the state of spiking models in computer vision applications.

The contributions of this work are summarized as follows:
\begin{enumerate}
    \item We develop a novel multi-scale spiking attention module, tailor-made for the SNN's attributes,  which enables the spiking transformer to extract features from different scales of inputs by spikes with low energy costs, improving the performance of the spiking transformer. 
    \item  {We develop a direct-training hierarchical spiking transformer, namely \modelname, incorporating MSSA into a vision transformer. This design marks the effective exploration of multi-scale spiking representation in Transformer-based SNNs.}
\item We conduct extensive experiments on mainstream static and neuromorphic datasets, achieving state-of-the-art performance compared to the latest SNN-based models. Notably, \modelname has surpassed QKFormer, which has achieved 84.22\% top-1 accuracy on ImageNet-1K with $224^2$ input size and 4-time steps using the direct training from scratch by  \hwrevised{85.06\%}, positioning itself as a state-of-the-art solution among SNN-transformer architectures.
\end{enumerate}
    Finally, based on the aforementioned experimental results, we conduct the ablation study to discuss and analyze \modelname. The source code is open-sourced and available at  {\url {https://github.com/Nanhu-AI-Lab/MSViT}}.







\section{Related Work}
\subsection{Vision Transformers}
ViTs segment images into patches and apply self-attention \cite{vaswani2017attention,devlin2018bert}  to learn contextual relationships, effectively reducing inductive bias \cite{neil2020transformers,neil2020transformers}  and outperforming CNNs across multiple vision tasks \cite{mei2021image,bertasius2021space,guo2021pct}. Nevertheless, ViTs face challenges like high parameter counts \cite{guo2021pct}, and increased computational complexity proportional to token length \cite{pan2020x,liu2022ecoformer}. To enhance the computational efficiency of ViTs, many researchers \cite{jie2023fact,li2023rethinking}  focused on exploring lightweight improvement methods from transformer architectures. For example, LeViT \cite{graham2021levit} incorporates convolutional elements to expedite processing, and MobileViT \cite{mehta2021mobilevit} combines lightweight MobileNet blocks with MHSA, achieving lightweight ViTs successfully. However, these enhancements still
rely on expensive MAC computations, which are not suitable for Edge devices. This highlights the need for investigating more energy-efficient ViT solutions. Involving SNNs in Transformer architectures is one of the approaches.

\subsection{Transformer Architecture in Spiking Neural Networks}

Transformer-based models have demonstrated remarkable capabilities in human-like text generation,  natural language understanding, and text-to-image \cite{vaswani2017attention,devlin2018bert}. 
As transformer-based networks have predominated in various tasks, researchers believe that the transformer architecture can also replicate the success when applied in spiking neural networks. 

Zhou et al. proposed a Spiking Transformer model, namely Spikformer, which models images into sparse visual features by using spike-form information without using a softmax operation for the first time \cite{zhou2022spikformer}. This form of spiking-transformer conducts bio-inspired spatio-temporal dynamics and spike (0/1) activations to obtain high energy efficiency from spiking self-attention. 
Subsequently, Zhou et al. proposed an SNN-based variant to improve Spikformer to Spikformer-v2 \cite{zhou2024spikformer}. 

The authors developed a Spiking Convolutional Stem (SCS) with supplementary convolutional layers and connected SCS to  Spikformer to enhance the image representation by spikes. Additionally, Spikformer-v2  introduced self-supervised learning for directly training SNNs.

Man et al. proposed a Spike-driven Transformer (SDT) \cite{yao2023spike},  which exploited only mask and addition operations without any multiplication and made up to 87.2$\times$ lower computation energy than vanilla self-attention. However, the experiment results in this work show that the top-1 accuracy on ImageNet-1K was only 77.1\% with 66.34 (M) parameters, which still has a large gap between ANN-based transformers. Therefore, the authors improved SDT to SDT-v2 \cite{yao2024spike-v2}, which extended the Spike-driven Transformer into a meta form. On ImageNet-1K, SDT-v2 achieved top-1 accuracy up to 80.0\%  with 55 (M) parameters, surpassing SDT-v1 by 3.7\%.
Zhou et al. proposed QKFormer \cite{zhou2024qkformer}, which enhanced the spiking transformer architecture by using a spiking Q-K attention module. Unlike the comment form of self-attention with Query (Q), Key (K), and Value (V), Q-K attention only adopts $Q$ and $K$ to implement the self-attention mechanism. The authors reported that QKFormer, significantly improved performance compared to Spikformer on ImageNet-1K.

\section{Methods}

\begin{figure}[!t]
    \centering
    \includegraphics[width=0.48\textwidth]{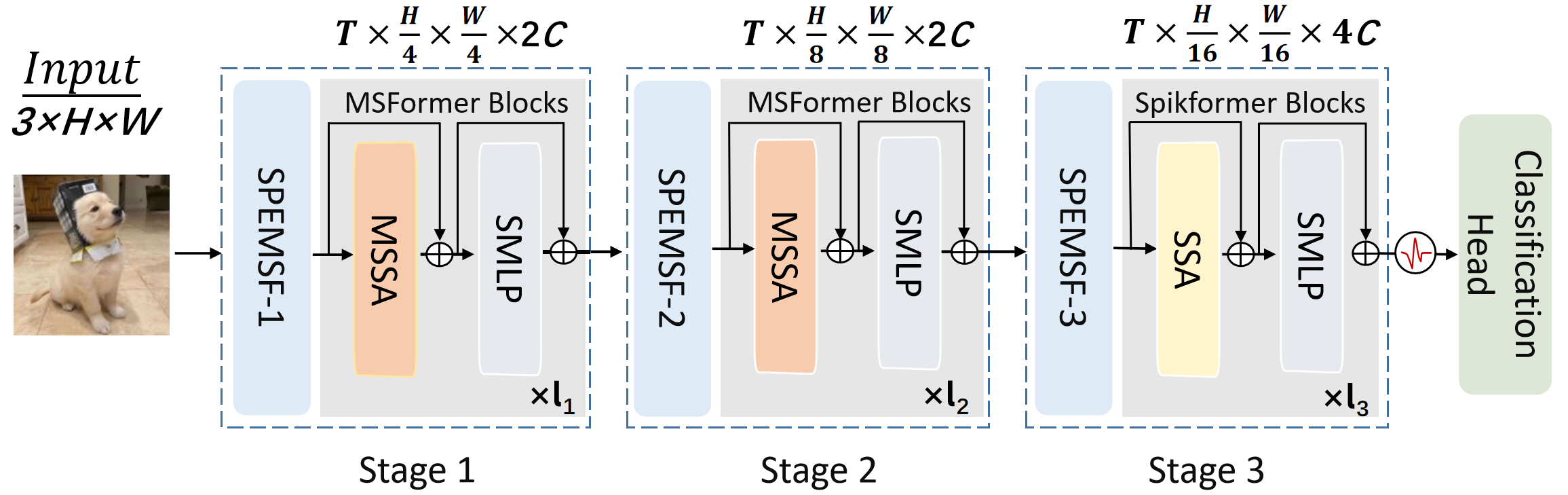}
    \caption{ Overview of \modelname, a hierarchical spiking transformer with multi-scale spiking attention. Note C denotes the spike-form dimension.}
    \label{fig:architecture}
\end{figure}

\begin{figure*}[!t]
    \centering
    \includegraphics[width=0.90\textwidth]{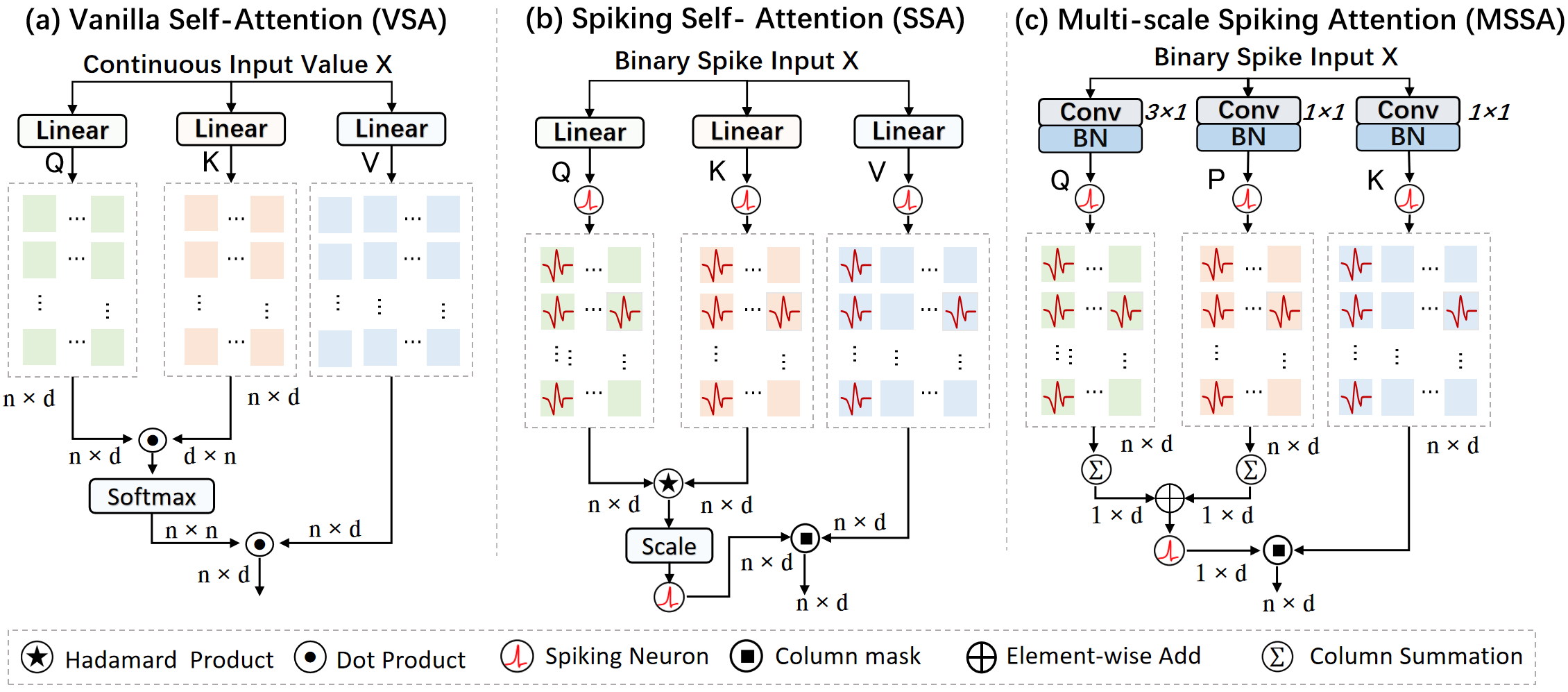}
    \caption{  Comparison of three self-attention computation paradigms. (a) VSA employs floating-point matrix multiplication to assess the spatial correlation between Q and K, resulting in a computational complexity of $ O(N^{2} D)$, (b) SSA lacks a dedicated temporal interaction module, maintaining
the same complexity as VSA and, (c)  MSSA introduces multi-scale interactions,
reducing the complexity to $O (T(ND)$).}
    \label{fig:MSSA}
\end{figure*}



In this section, we first present the overall architecture of \modelname. Secondly, we introduce the important components of \modelname, including hybrid spiking attention integration, Spiking Patch Embedding with Multi-scale Feature Fusion (SPEMSF) which serves as the tokenization method for \modelname, and MSFormer Blocks. Finally, we introduce MSSA and SSA in detail.  

\subsection{Model Architecture}

\subsubsection{Overall of Architecture}
The overview of \modelname is illustrated in Figure \ref{fig:architecture}. The input $I$ of \modelname is represented as $T \times C_0 \times H \times W$, where $T$ denotes the timesteps, $C_0$ denotes the number of channels, and $H$ and $W$ denote the height and width of the inputs, respectively. 

When training \modelname on static RGB image datasets, $T$ is  $1$ and $C_0 $ is set to $3$. For neuromorphic datasets, $T = T_0$, $C_0 = 2$. In the encoder, A patch size of $4 \times 4$ is used, and the input feature $(4 \times 4 \times C_0)$ is transformed into a spike-form representation with $C$ channels (note $C$ distinguishes from $C_0$) using Spiking Patch Embedding with Multi-scale Feature Fusion-1 (SPEMSF-1).

Given an input $I \in \mathbb{R}^{T \times C_0 \times H \times W}$, SPEMSF-1 first transforms $I$ into a series of tokens $x_n$ ($n \in N$), where $x \in \mathbb{R}^{T \times H \times W \times C}$ and $N = \frac{H}{4} \times \frac{W}{4}$. Along with the subsequent MSFormer block, this constitutes "Stage 1". To construct a hierarchical spiking transformer, the number of tokens $n$ is further reduced in SPEMSF-2 and SPEMSF-3 to $ \frac{H}{8} \times \frac{W}{8}$ and $ \frac{H}{16} \times \frac{W}{16}$, respectively. These transformations are handled in "Stage 2" and "Stage 3", where each stage reduces the token dimensions by a $2 \times 2$ patch.

In the first and the second stages, the number of channels $C$ is set to 2, while it is increased to 4 in Stage 3. The number of spiking transformer layers $L$ in each stage is configured as $l_1$, $l_2$, and $l_3$, based on the model size and the dataset being trained (detailed settings are provided in Section \ref{experiments}). 

Together, the three stages collectively implement the hierarchical spiking-transformer architecture for \modelname.

\subsubsection{Hybrid Spiking Attention Block Integration}

While Spiking Neural Networks (SNNs) are inherently energy-efficient, the amount of information that can be processed through spikes in transformer architectures remains limited compared to float values \cite{zhang2022spiking,zhou2022spikformer}, resulting in suboptimal performance.

To address this limitation, we integrate multi-scale spiking attention (MSSA) into \modelname. However, applying a single type of MSSA across all stages of \modelname leads to an increase in model size, which is not ideal. To optimize the trade-off between performance and efficiency, we revisit the model design and propose a hybrid spiking attention mechanism for \modelname.

In the first and second stages, we adopt MSSA within the MSFormer blocks to enhance feature extraction. In the final stage, we employ standard spiking self-attention (SSA), as used in Spikformer \cite{zhou2022spikformer}, to process the deeper layers of the model. This hybrid design strikes a balance between model size and performance, enabling \modelname to achieve improved accuracy while maintaining a relatively low energy cost during training.

\subsubsection{MSFormer Blocks}

Meta-former \cite{yu2022metaformer} established that the general structure of transformers can be characterized by two key components: a token mixer and a channel mixer. In conventional transformer architectures, the token mixer is often implemented by an attention block, while an MLP block implements the channel mixer.

Similarly, \modelname adopts this structure and is referred to as MSFormer. Each MSFormer block comprises two modules: a multi-scale spiking attention (MSSA) module and a Spiking MLP (SMLP) module. The formulation of the block can be expressed as follows:
 \begin{equation}
X'_l=MSSA(X_{l-1})+X_{l-1}, X'_{l}\in \mathbb{R}^{T\times N \times D}, 
\end{equation}
\begin{equation}
X_l=SMLP(X_{l-1})+X_{l-1}, X_{l}\in \mathbb{R}^{T\times N \times D}, 
\end{equation}
where $N$ is the length of patches and $D$ is the embedding size.

\subsubsection{Spiking Patch Embedding with Multi-scale Feature Fusion }

In this section, we describe the proposed Spiking Patch Embedding with Multi-Scale Feature Fusion (SPEMSF) in detail. For vision tasks, higher-level feature maps often preserve rich semantic information but suffer from lower resolution, whereas lower-level feature maps capture raw input details with higher resolution. To address this trade-off, some ANN-based methods \cite{chen2018encoder,badrinarayanan2017segnet} combine feature maps across layers. However, the improvements in these works are limited due to the semantic gap between feature maps at different scales.

Inspired by these works, we propose a multi-scale feature fusion approach to enhance the spike representation in patch embedding for \modelname. Our approach leverages convolutional layers with different kernel sizes to process input features at varying semantic levels, effectively treating them as multi-scale features. The fused features, represented as spikes, preserve both low-level and high-level semantic information, reducing information loss during embedding.

The core idea is to use lightweight convolutional operations to project input spiking maps into multiple feature channels. These projections are fused to generate a richer representation for subsequent processing. Specifically, we utilize a combination of linear projections $W_d$ to achieve this.

Given the input spiking map $X$, the patch embedding process is formulated as: \begin{equation} Y = \mathcal{F}(X, W_i) \oplus \mathcal{G}(X, W_j), \end{equation} where $\mathcal{F}$ and $\mathcal{G}$ are functions representing different multi-scale transformations, and $\oplus$ denotes the fusion operation.

In our implementation:
\begin{itemize}
    \item The linear projection $W_d$ in function $\mathcal{F}$ is defined as a lightweight convolutional layer with a $1 \times 1$ kernel and stride $\geq 1$, which focuses on channel-wise transformations,
    \item function $\mathcal{G}$ uses a $3 \times 3$ convolutional layer with stride $=2$, incorporating more spatial context while reducing resolution.
\end{itemize}

The implementation for function $\mathcal{F}$ is a simple pipeline of {Conv2D-BN-SNN}, while $\mathcal{G}$ is designed with one of the following configurations: (1) {Conv2D-BN-MaxPooling-SN-Conv2D-BN-SNN}, or (2) {Conv2D-BN-SN-Conv2D-BN-MaxPooling-SNN}.
This multi-scale feature fusion not only enriches the spike representation for inputs but also ensures compatibility with the channel and token requirements of the patch embedding block. The figure of Spiking Patch Embedding is further illustrated in \hwrevised{Appendix B}.

\subsection{Multi-scale Spiking Attention (MSSA)}

We propose a novel Spike-driven Transformer that maintains the spike-driven nature of Spiking Neural Networks (SNNs) throughout the network while achieving strong task performance by incorporating multi-scale features into the attention module.

The overview of Multi-Scale Spiking Attention (MSSA) is presented in Figure \ref{fig:MSSA} (c). For comparison, traditional Vanilla Self-Attention (VSA) and Spiking Self-Attention (SSA) which is the core component of Spikformer, are shown in Figure. \ref{fig:MSSA} (a) and (b). Both VSA and SSA use three components (\(Q, K, V\)) and have computational complexity of \(O(N^2d)\) or \(O(Nd^2)\). In contrast, MSSA achieves linear complexity of \(O(Nd)\). The initialization of the three components in MSSA is defined as follows:

\begin{align}
 Q = \mathcal{SN}(BN(XW_Q)), \\ P = \mathcal{SN}(BN(XW_P)), \\V = \mathcal{SN}(BN(XW_V)),
\end{align}

where \(X \in \{0,1\}^{T \times N \times D}\) is the spiking map of the input, \(N\) represents the number of patches, and \(D\) is the feature dimension. \(Q\) and \(P\) represent lower and higher-level spiking feature maps of \(X\), generated by convolutional layers. All three components (\(Q, P, V\)) are produced through learnable linear matrices. Here, \(\mathcal{SN}\) denotes the spiking neuron layer, and \(BN\) represents the batch normalization layer.

Unlike VSA (a) and SSA (b), MSSA replaces matrix multiplication with column summation to compute the attention interaction between the components (\(Q, P, V\)). The attention mechanism in MSSA is defined as:
\begin{equation}
\text{MSSA}(Q, P, V) = \mathcal{SN}(\text{SUM}_c(Q) \oplus \text{SUM}_c(P)) \otimes V,
\end{equation}
where \(\oplus\) represents element-wise addition, and \(\text{SUM}_c(\cdot)\) performs column-wise summation, resulting in an \(N \times 1\) vector \(\alpha\), which encodes self-attention scores in spike form:
\begin{equation}
\alpha_q = \sum_{i=0}^d Q_i, \quad \alpha_p = \sum_{i=0}^d P_i.
\end{equation}

To implement multi-scale feature fusion, \(\alpha_q\) and \(\alpha_p\) are mixed by element-wise addition (\(\oplus\)), fusing the lower-level feature \(Q\) with the higher-level  feature  \(P\) via the attention scores. The resulting spike-based attention vector is then applied to the \(V\) spike matrix to generate the final spike representation of \(X\) using MSSA.

 \subsection{Spiking Self Attention}

Spikformer \cite{zhou2022spikformer} introduced a novel spike-based self-attention mechanism, termed Spiking Self-Attention (SSA). Unlike traditional self-attention, SSA leverages sparse spike-form representations (\(Q, K, V\)) and eliminates the need for softmax operations and floating-point matrix multiplications. The computational process of SSA is formulated as:

\begin{align}
  \quad  Q = \mathcal{SN}(BN(XW_Q)),\\  K = \mathcal{SN}(BN(XW_K)),\\V = \mathcal{SN}(BN(XW_V)),
  \end{align}
\begin{equation}
SSA(Q, K, V) = \mathcal{SN}((QK^{T}) \otimes V \cdot s),
\end{equation}

where \(Q, K, V \in \mathbb{R}^{T \times N \times D}\) are spike-form representations computed through learnable linear transformations. Here, \(s\) is a scaling factor. \(\mathcal{SN}\) denotes the spiking neuron layer and \(BN\) represents the batch normalization layer.

In stage 3 of \modelname, we adopt SSA to perform spiking attention in the deeper layers, leveraging its efficiency and alignment with the spike-driven computation paradigm.

\section{Experiments \label{experiments}}

\begin{table*}
    \centering
    \begin{tabular}{ccccccccc}
        \midrule
        Method &spiking & Architecture& \makecell{Params\\ (M)} &\makecell{ Input\\ Size} & \makecell{Time \\ Step}  & \makecell{Energy\\(mJ)} & \makecell{Top-1 Acc.\\(\%)} \\
        \midrule

        DeiT    \cite{touvron2021training}  &    \XSolidBrush      &DeiT-B   & 86.60 & $224^2$ & 1 & 80.50 & 81.80  \\
        VIT-B/16 \cite{dosovitskiy2020image} & \XSolidBrush          & ViT-12-768   & 86.59 & $384^2$ &1 & 254.84 & 77.90  \\
     \multirow{2}{*}{Swin Transformer \cite{liu2021swin}}      & \checkmark         &  Swin-T   & 28.50 & $224^2$ & 1 &70.84 & 81.35  \\
                                                             & \XSolidBrush         &  Swin-S   & 51.00 & $224^2$ & 1 & 216.20 & 83.03  \\
                \hline
       \multirow{2}{*}{ Spikformer \cite{zhou2022spikformer} }   & \checkmark          & 8-384  & 16.80 & $224^2$ &4 & 5.97 &  70.24  \\
           & \checkmark          & 8-768 & 66.30 & $224^2$ &4 & 20.0 & 74.81  \\
           \cline{3-8}
        \multirow{2}{*}{Spikformer V2 \cite{zhou2024spikformer}}    & \checkmark            & V2-8-384& 29.11 & $224^2$ &4 &  4.69 & 78.80  \\
           & \checkmark    & V2-8-512 & 51.55 & $224^2$ & 4 &  9.36 & 80.38  \\
        \midrule
       \multirow{3}{*}{Spike-driven \cite{yao2022attention}}   & \checkmark        &  SDT 8-384    & 16.81 & $224^2$ & 4 &  3.90 & 72.28  \\
                                                                    & \checkmark    &  SDT8-512  & 29.68 & $224^2$ & 4 &  4.50 & 74.57  \\
                                                                    & \checkmark    &  SDT8-768  & 66.34 & $224^2$ & 4 &  6.09 & 77.07  \\
                                                                    \cline{3-8}
        \multirow{3}{*}{Spike-driven v2 \cite{yao2024spike-v2}}       
        & \checkmark   &  SDT v2-10-384 & 15.10 & $224^2$ & 4 & 16.70 & 74.10 \\
        & \checkmark       &   SDT v2-10-512    & 31.30 & $224^2$ & 4 &  32.80 & 77.20  \\
                  & \checkmark   &  SDT v2-10-768 & 55.40 & $224^2$ & 4 & 52.40 & 80.00 \\
                   
        \midrule
        \multirow{3}{*}{QKFormer \cite{zhou2024qkformer}}          & \checkmark     & QK-10-384  & 16.47 & $224^2$ &4 & 15.13 & 78.80  \\
              & \checkmark       &QK-10-512  & 29.08 & $224^2$ & 4 & 21.99 & 82.04  \\
              & \checkmark        &QK-10-768  & 64.96 & $224^2$& 4 & 38.91 & 84.22  \\
        \midrule
        \multirow{3}{*}{\modelname}      & \checkmark    & \modelname-10-384 & 17.69 & $224^2$ & 4 & 16.65 &\textbf{80.09}  \\ 
        
            & \checkmark   &\modelname-10-512 & 30.23 & $224^2$ &4 & 24.74 & \textbf{82.96}  \\
              & \checkmark    &\modelname-10-768 & 69.80 & $224^2$ &4 & 45.88 & \textbf{85.06}  \\

        \hline
    \end{tabular}
    \caption{  Comparison of \modelname performance to respective ANN-based and SNN-based state-of-the-art models by accuracy (Acc.\%). The experimental results are on ImageNet-1K. Energy is calculated as the average theoretical power consumption
when predicting an image from ImageNet test set. The energy data for \modelname and ANNs is
evaluated according to Appendix C.}
    \label{tab:imagenet_classfication}
\end{table*}

\begin{table*}[h]
\begin{tabular}{ccccccccccccc}
\hline
\multirow{2}{*}{method} & \multicolumn{3}{c}{CIFAR10} & \multicolumn{3}{c}{CIFAR100} & \multicolumn{3}{c}{DVS128} & \multicolumn{3}{c}{CIFAR10-DVS} \\
\cline{2-4} \cline{5-7}  \cline{8-10} \cline{11-13} 

                        & Param     & T    & Acc      & Param     & T     & Acc      & Param    & T     & Acc     & Param      & T       & Acc      \\

                        \midrule
Spikformer              & 9.32      & 4    & 95.51    & 9.32      & 4     & 78.21    &    2.57      & 16    & 98.3    &    2.57        & 16      &    80.9      \\
SDT \cite{yao2023spike}& 9.32      & 4    & 95.81    & 9.32      & 4     & 78.21    &      2.57    & 16    & 99.3   &      2.57      & 16      &    80.9      \\
CML \cite{wang2023spatial}& 9.32      & 4    & 95.81    & 9.32      & 4     & 80.02    &    2.57      & 16    & 98.6    &    2.57        & 16      &     80.9     \\
QKFormer   \cite{zhou2024qkformer}       & 6.74      & 4    & 96.08     &      6.74     & 4     &     81.12     &     1.50     &   16    &  98.6       &     1.50       & 16      &    84.0      \\
\midrule
ResNet-19           & 12.63      & -    & {94.97}    & 12.63     & -     & 75.35    &    -      & -   & -    &       -     & -      &     -     \\
Transfomer (4-384)           & 9.32      & -    & \textbf{96.73}     & 9.32      & -    & 81.02   &    -    &   -    & -        &    -        & -      &   -       \\
\midrule
\modelname           &   7.59        & 4    &     {96.53}     &      7.59     & 4     &      \textbf{81.98}    &    1.67      &16     &      98.80   &       1.67     & 16      &\textbf{84.30} \\
\hline
\end{tabular}
\caption{Comparision on CIFAR10, CIFAR100, DVS128, CIFAR10-DVS. "Param" denotes "Parameter (M)", "Acc" denotes "Top-1 Accuracy (\%)", "T" denotes "Time Step".}
\label{tab:cifar_classifcation}
\end{table*}

This section introduces the details of the experiment, including data collection, implementation, and evaluation methods.


\subsection{Experimental Setup}

We evaluate \modelname on both static image classification and neuromorphic classification tasks. For static image classification, we use ImageNet-1K \cite{deng2009imagenet} and CIFAR10/100 \cite{krizhevsky2009learning}. For neuromorphic classification, we employ the CIFAR10-DVS \cite{li2017cifar10} and DVS128 Gesture \cite{amir2017low} datasets. Appendix D introduces the datasets in detail.

\subsection{Results on ImageNet-1K Classification }
\textbf{Experimental Setup on ImageNet-1K.}
We adopt a training recipe similar to that proposed in \cite{zhou2024qkformer} and detail the configurations in this section. First, the model is trained in a distributed manner for 200 epochs on an 8-A100 GPU server. We employ several data augmentation techniques, including RandAugment \cite{cubuk2020randaugment}, random erasing \cite{zhong2020random}, and stochastic depth \cite{huang2016deep}, with a batch size of 512. Additionally, gradient accumulation is utilized to stabilize training, as suggested in \cite{he2022masked}. Second, the optimization process leverages synchronized AdamW with a base learning rate of $6 \times 10^{-4}$ per batch size of 512. The learning rate is linearly warmed up at the initial stage and subsequently decays following a half-period cosine schedule. The effective runtime learning rate is scaled proportionally to the batch size, calculated as \texttt{BatchSize/256} multiplied by the base learning rate. Finally, the architecture is designed with three stages (as illustrated in Figure \ref{fig:architecture}), where the number of layers in each stage is configured as $\{l_1=1, l_2=2, l_3=7\}$, respectively. These configurations collectively ensure robust and efficient training of the proposed model.

\noindent\textbf{Primary Results on ImageNet-1K.}
\hwrevised{ The experimental results demonstrate the superior performance of our
proposed \modelname, surpassing previous works’ performance. Generally, \modelname (69.80 M) with input size of $224^2$ achieves 85.06\% top-1 accuracy and 97.58\% top-5 accuracy on ImageNet-1K, which performs the best in Table \ref{tab:imagenet_classfication}. To start with the experiments, we first compare \modelname with Spikformer which is the first version of the spike-form transformer\cite{yao2022attention}.  Our \modelname (69.80 M, 85.06\%) significantly outperforms Spikformer (66.34 M, 74.81\%, by 10.25\% with the input size of $224{^2}$. In Table \ref{tab:imagenet_classfication}, we can see that \modelname achieves the best performance on accuracy, utilizing slightly more parameters. Additionally, compared to SDSA, MSSA has lower computational complexity.
Meanwhile, \modelname outperforms Spike-driven Transformer \cite{yao2023spike} (SDT, built by SDSA) by 7.81\%, 8.39\%, and 8.58\% respectively at three model size levels (17.69M, 30.23M, 69.80M), and surpasses QKFormer by 1.29\%, 0.92\%, and  0.84 \%  by comparable parameters.  
Surprisingly, \modelname obtains significant improvement,  outperforming  MST-T \cite{wang2023masked} by 4.45\% with 30.23M parameters, and 45.88 mJ energy cost. }

\noindent \textbf{Comparing with ANN Models on ImageNet.}
 \modelname is designed as an event-driven SNN model, in which the outputs of the embedding layers, the matrix calculation in the attention blocks, and information transmission are binary spikes $\{0,1\}$. As a result, the 
 multiplications of the weight matrix and activations, such as RELUs, which are important, can
be replaced by AND-ACcumulate (AC) operations, which benefit model training with high energy efficiency. Meanwhile, to achieve a competitive performance for ANN-based models, we apply the hierarchical transformer architecture to \modelname. Although the implementation of hierarchical architectures is relatively more complex than those using the same attention blocks throughout the networks (\cite{zhou2022spikformer,yao2023spike}), it is still worth using a hierarchical architecture to reduce the performance gap between ANNs and SNNs. This is mainly because the hierarchical architecture naturally has the flexibility to model at various scales and has linear computational complexity with respect to image input size \cite{liu2021swin}. For instance, our \modelname outperforms the most well-known Transformer-based ANNs in performance with high energy efficiency under the same experiment conditions without pretraining or extra training data,  among \modelname (69.80M, \hwrevised{85.06\%}, SNN, \hwrevised{45.88}mJ), Swin Transformer (88M, 84.5\%, ANN, 216.20mJ) \cite{liu2021swin}, DeiT-B (86M, 83.1\%, ANN, 254.84mJ) \cite{touvron2021training}
and ViT (85.59M, 77.9\%, ANN, 254.84mJ) \cite{dosovitskiy2020image} ( the detailed results are in Table \ref{tab:imagenet_classfication}).

\subsection{Results on CIFAR and Neuromorphic Datasets} \label{sec:cifar100}
\textbf{CIFAR Classification.}
We conduct experiments on smaller datasets and configure the training process to ensure sufficient model optimization. Specifically, we train the model for 400 epochs with a batch size of 128, following the setup of Spikformer \cite{zhou2022spikformer}. For the three-stage training of \modelname, we utilize a total of 4 blocks distributed as $\{1, 1, 2\}$ across the stages. Thanks to the hierarchical architectural design, \modelname comprises 7.59M parameters, which is slightly larger than QKFormer (6.74M) but smaller than Spikformer (9.32M). The performance results on the CIFAR datasets are summarized in Table \ref{tab:cifar_classifcation}. On CIFAR10, \modelname achieves an accuracy of \textbf{96.53\%}, outperforming Spikformer by 1.02\% and QKFormer \cite{zhou2024qkformer} by 0.35\%. For CIFAR100, \modelname achieves an accuracy of \textbf{81.98\%}, exceeding Spikformer (78.21\%) by \textbf{3.77\%} and QKFormer by 0.86\%. Notably, \modelname surpasses Vision Transformer (ViT), an ANN-based model, on CIFAR100 by 0.93\%. This represents a relatively significant improvement over other SNN-based Transformer architectures, demonstrating the efficacy of \modelname in classification tasks on various datasets.

\noindent  \textbf{Neuromorphic Classification.}
To evaluate \modelname on neuromorphic tasks, we compare our model with the state-of-the-art models using the CIFAR10-DVS and DVS-Gesture datasets. Unlike conventional static image datasets, neuromorphic datasets comprise event streams instead of RGB images. This introduces a significant domain shift between the source (static image datasets) and target (neuromorphic datasets) domains for models pre-trained on static images. We aggregate events over specific time intervals to address this discrepancy to construct frames. The RGB channels are replaced with positive, negative, and the sum of events as input features, respectively.
For this experiment, we implement a lightweight version of \modelname with only 1.67M parameters, utilizing a block configuration of $\{0, 1, 1\}$ across the three stages. The maximum patch embedding dimension is set to 256. The model is trained for 200 epochs on the DVS128-Gesture dataset and 106 epochs on the CIFAR10-DVS dataset. The number of time steps for the spiking neurons is set to either 10 or 16. 
The experimental results for temporal neuromorphic classification are summarized in Table \ref{tab:cifar_classifcation}. On the DVS128-Gesture dataset, \modelname with 1.67M parameters achieves an accuracy of 98.80\% using 16-time steps and 98.37\% using 10-time steps. For the CIFAR10-DVS dataset, \modelname achieves an accuracy of 84.30\% using 16-time steps, significantly outperforming Spikformer by 3.4\%. Moreover, with 10-time steps, \modelname achieves an accuracy of 83.80\%, surpassing Spikformer by 4.20\% and QKFormer by 0.3\%. These results highlight the efficiency and performance gains of \modelname, with a minimal parameter count.


\section{Ablation Study} \label{sec:ablation}

\begin{table}
    \centering
    \begin{tabular}{lp{1.3cm}<{\centering}p{1.3cm}<{\centering}}
        \midrule
        Model   & Param (M) & CIFAR100 (Acc) \\ 
        \midrule
       
        MSSA (P+P)  +SSA           & 7.74 & 81.36  \\
        MSSA (Q+Q)  +SSA           & 7.45 & 81.56  \\
        MSSA (P)  +SSA           & 7.52 & \hwrevised{81.35} \\
        MSSA(Q)   +SSA          & 7.37 &   \hwrevised{81.44} \\
        \midrule
                MSSA  +MSSA(P+P)           & 9.04 & 81.25 \\
        MSSA +MSSA (Q+Q)          & 7.89  & 81.15 \\
                MSSA  +MSSA (P+Q)          & 8.48 & {81.65} \\
        \midrule
         MSSA(P+Q) + SSA (\modelname )           & 7.59 & \textbf{81.98}  \\
        \hline
    \end{tabular}
    \caption{Ablation study of MSSA with different feature fusion dimensions. P: feature from $3 \times 1$ conv; Q: feature from $1 \times 1$ conv.
    }
    \label{tab:ablation_studies}
\end{table}

{
\textbf{{Hybrid Spiking Attention Integration.}}
We test \modelname on CIFAR100 and use the \modelname equipped with MSSA (on stage 1,2) and SSA (on stage 3) as the baseline.  The results show that using the same MSSA at each stage achieves relatively high performance with 81.65\% on Top-1 accuracy among the study cases in Table \ref{tab:ablation_studies}. However, the number of parameters of \modelname increases too much, which may incur Unworthy computational consumption, especially in the case we conduct experiments on large-scale datasets, such as the ImageNet-1K. This is mainly because a ($3\times 3$) kernel in convolution layers consumes more parameters than a point-wise convolution layer (with ($1\times 1$) kernels) to extract features from a larger \hwrevised{perception field}. Most of the baselines, including Spikformer \cite{zhou2022spikformer}, QKFormer \cite{zhou2024qkformer}, and SDT \cite{yao2024spike-v2}, solely use the \underline{P}oint\underline{W}ise \underline{Conv}olution layer (PWConv) to conduct information extractions. Although  PWConv decreases the number of parameters throughout the model, it may cause information loss of high-level features at the shallow layers of spike-form transformer architectures, which are stage 1 and stage 2 in our \modelname. Particularly, using the binary spikes that only use $\{0,1\}$ to transfer information is difficult. We addressed this issue and conducted the fusion of the low-level feature $Q$, and the high-level feature  $P$ to transmit more features to the model. Finally, we adopted the hybrid spiking attentions (MSSA + SSA) as the final version of \modelname to strike the trade-off between the computational efficiency and performance of the model. \hwrevised{The experimental results also show that \modelname gains great performance with 81.98\% on CIFAR100 using only 7.59M parameters.}

\noindent\textbf{The effectiveness of MSSA.}
 Lines 1- 4 in Table \ref{tab:ablation_studies} illustrate the effectiveness of feature fusion at stages 1 and 2 of \modelname. Line 1 shows that MSSA (P+P) performs 81.25, which is almost the same as the result of  MSSA (P) (Line 3). \hwrevised{This indicates that the fusion for the same feature sizes can not improve its performance for \modelname.} Line 2 shows MSSA(Q+Q) obtains only 81.15, which is similar to QKformer's ( only using $1\times 1$ kernels ) experimental result. In conclusion, adopting only low-level/local feature maps may still incur semantic information loss, damaging the spiking representation of input images. 
}



\section{Conclusion}
In this work, we design a novel spike-driven multi-scale attention (MSSA), which involves a multi-scale feature fusion mechanism in the spiking hierarchical transformer architecture to improve the model's performance. MSSA fuses low-level and high-level information of inputs, significantly improving the model performance with limited parameter increases. Furthermore, MSSA replaces the dot-product or Hadamard product existing in VSA or SSA with a column sum, maintaining the computation in linear complexity to the tokens of the inputs. Correspondingly, we also utilize Spiking Patch Embedding with Multi-scale Feature Fusion (SPEMSF), which enhances spiking representation for both high-level and low-level information of inputs, thereby promoting model improvement. Finally, we implement a hierarchical spiking transformer architecture equipped with the aforementioned MMSA and SPEMSF, namely \modelname. We have conducted extensive experiments, and the results show that our model achieves state-of-the-art performance on both static and neuromorphic datasets. Notably, \modelname achieved top-1 accuracy on ImageNet-1K over 85\% with 69.80M parameters and image input of size $224^2$ by direct training from scratch. Leveraging the \modelname's superior capabilities, we strive to inspire confidence in the application of Spiking Neural Networks through our work.

\appendix
\section{Spiking Neurons}
In this work, we use the Leaky Integrate-and-File(LIF) model as the spiking neurons in our model. The dataflows of a LIF neuron are formulated as follows:
\begin{equation}
H_{[t]}=V_{[t-1]}+\frac{1}{\tau}(X_{[t]}-(V_{[t-1]}-V_{reset}),
\label{eq:lif}
\end{equation}

\begin{equation}
S_{[t]}=\theta(H_{[t]}-V_{th}),
\label{eq:spiking}
\end{equation}

\begin{equation}
V_{[t]}=H_{[t]}(1-S_{[t]})+V_{reset}S_{[t]},
\label{eq:mem}
\end{equation}
where $\tau$ is the membrane time constant, $X_{[t+1]}$ is the input current at time step $t+1$, $V_{reset}$ is the reset potential, and $V_{th}$ is the threshold of the spike firing.  Once the membrane potential $H_{[t]}$ reaches the firing threshold $V_{th}$, the spiking neuron generates a spike $S_{[t]}$.$\theta$ is the Heaviside step function, which can be formulated as:
\begin{equation}
    \theta(V)=
 \begin{cases}
1, & \quad \geq V_{th}, \\
0, & \quad otherwise,
\end{cases}
\label{eq:snn_spiking_function}
\end{equation}

$V_{[t]}$ is the membrane potential in a LIF neuron. It remains $H_{[t]}$ when the potential is under the $V_{th}$. Once the neuron fires, $V_{[t]}$ is reset as $V_{reset}$.

\begin{figure}
    \centering
    \includegraphics[width=0.5\textwidth]{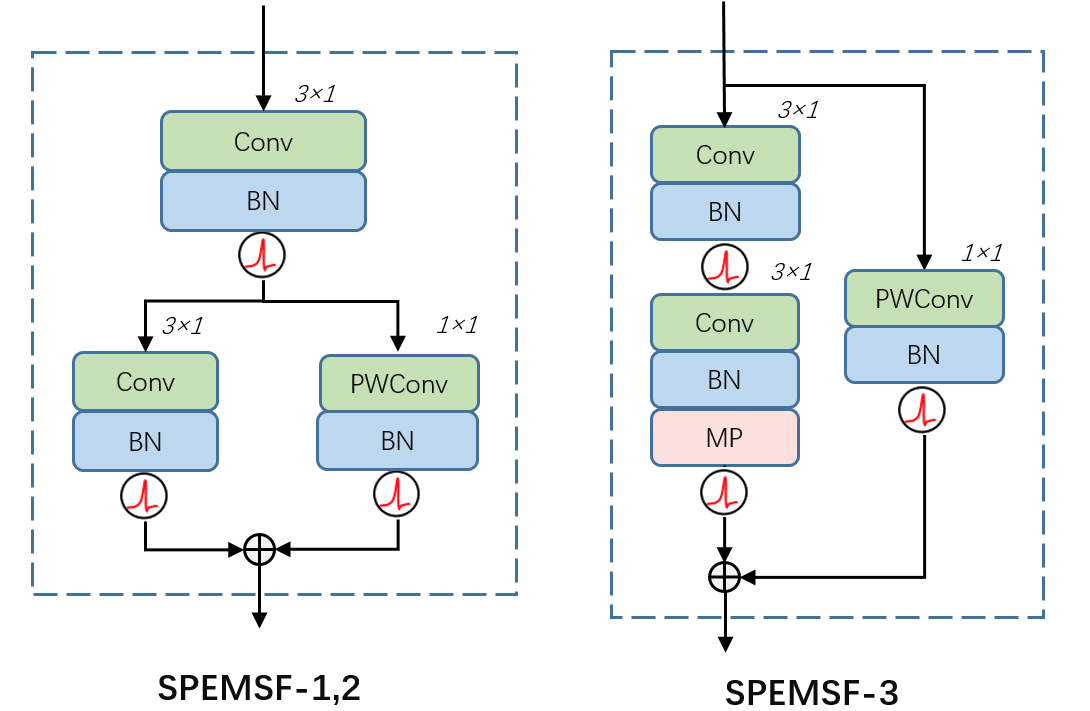}
    \caption*{\textbf{Appendix Figure 1}. Overview of SPEMSF, a hierarchical spiking transformer with multi-scale spiking attention. Note C denotes the spike-form dimension.}
    \label{fig:SPEMSF}
\end{figure}

\section{The structure of Spiking Patch Embedding with Multi-scale Feature Fusion}
Section 3.1 introduces the model's architecture with Spiking Patch Embedding and Multi-scale Feature Fusion. We provide the structure of SPEMSF in Appendix Figure 1.

\section{Energy Consumption Calculation on ImageNet}

When calculating the theoretical energy consumption, the consumption of BN layers could
be ignored. We calculate the number of Synaptic Operations (SOPs) of the spike before calculating
theoretical energy consumption for our MSViT.
\begin{equation}
    SOP^l = fr \times T \times FLOP_s^l
    \label{eq:sop}
\end{equation}

where $l$ is the layer/block number in \modelname, $fr$ is the firing rate of the layer/block,  and $T$ is the simulation time step of spiking neurons. $FLOP_s^l$ represents the floating point operations of a layer in MSViT, which are the multiply-and-accumulate (MAC) operations. $SOP^s$ is the amount of spiking operations, that is, accumulate (AC) operations. We assume that the MAC and AC operations are executed on a 45nm hardware \cite{horowitz20141,zhou2022spikformer,shi2024spikingresformer}, where $E_{MAC} = 4.6pJ$ and
$E_{AC} = 0.9pJ$. The theoretical energy consumption of \modelname therefore can be calculated as follows:
\begin{equation}
\begin{multlined}
E=E_{AC}\times (\sum_{l_1=2}^{L_1} SOP^{l_1} _{Conv}+\sum_{l_2=1}^{L_2} SOP^{l_2} _{MSSA} +\\\sum_{l_3=1}^{L3} SOP^{l_3}_{SSA})+EMAC\times(FLOP^{1}_{Conv})
\end{multlined}
\label{eq:enery}
\end{equation}
Equation \ref{eq:enery} shows the calculation method of energy consumption for \modelname. $FLOP^{1}_{Conv}$ is the entrance layer that encodes floating-point input into spike-form for MSSAs and SSAs.  $L$ represents the number of layers of \modelname. The distribution of $L$ is $L_1,L_2,L_3$, which is $\{1,2,7\}$, corresponding to the number of MSSA and SSA layers at stage 1, stage, 2, and stage 3 based on our architecture design. After all, we sum up all the operations and multiply them by $E_{AC}$.

As for ANN-based transformer architectures, the theoretical energy consumption can be calculated as:

\begin{equation}
E_{MAC} \times FLOP_s
    \label{Eann}
\end{equation}

\section{Dataset Information}
To verify the effectiveness of our proposed \modelname, we evaluated the model on mainstream static and neuromorphic datasets.
 \textbf{Static Image Datasets}. ImageNet-1K is a widely recognized benchmark for classification tasks, comprising 1.28 million training images and 50,000 validation images across 1,000 categories. CIFAR10 and CIFAR100 provide a total of 50,000 training images and 10,000 test images, each with a resolution of $32 \times 32$. The key difference between these datasets lies in their number of categories: CIFAR10 includes 10 classes, while CIFAR100 encompasses 100 classes, presenting a more challenging classification task due to finer-grained distinctions.

\textbf{Neuromorphic Datasets.} CIFAR10-DVS is an event-based neuromorphic dataset derived from the CIFAR10 static image dataset, generated by capturing shifting image samples with a Dynamic Vision Sensor (DVS) camera. It includes 9,000 training samples and 1,000 test samples. The DVS128 Gesture dataset is a gesture recognition benchmark containing 11 hand gesture categories performed by 29 individuals under three different illumination conditions. This dataset captures event-based neuromorphic data, making it well-suited for testing models on dynamic tasks.
These datasets collectively provide a comprehensive evaluation framework for assessing the performance of \modelname across both static and event-based classification tasks.

\section{Limitation}
\textbf{Experimental settings.}
In the first and the second stages, the number of channels $C$ is set to 2, while it is increased to 4 in Stage 3.
This is because the model is expected to receive sufficient information via enough channels at the shallow stages (1-2), avoiding information loss. However, this setting increases the parameters. This issue will be improved in future works.

\textbf{Further analysis}. It is necessary to clarify how multi-scale processing influences the differential responses of spiking neurons. To address this, we compare the firing rates between MSViT and QKFormer, and observe that the firing rate of MSViT is 0.306, which is higher than 0.285 observed in QKFormer. This suggests that the spiking neurons in MSViT exhibit greater activity and transmit more information compared to those in QKFormer. This issue will be improved in future works.


\textbf{Evaluation on more tasks}. Our current model is primarily designed for image and DVS classification tasks. To explore its broader potential, we plan to extend its application to additional tasks, such as segmentation, detection, and particularly language tasks, to demonstrate its generalizability. Furthermore, we intend to develop more efficient and high-performance network architectures that require fewer time steps, incorporating multi-scale attention mechanisms and other efficient modules. This work will help us further reduce the computational resources required for edge-device training.

\bibliographystyle{named}
\bibliography{ijcai25}

\end{document}